\documentclass{article}

\usepackage{graphicx} 
\usepackage{color}
\usepackage{authblk}
\usepackage{appendix}
\usepackage{cite}
\usepackage{url}
\usepackage{ascmac}

\usepackage[legacycolonsymbols]{mathtools}
\usepackage{amsthm,amsmath,amsfonts,amssymb}
\usepackage{mathabx}
\usepackage{algorithm,algorithmic}
\usepackage{bm}
\usepackage{bbm}
\usepackage{array}
\usepackage{booktabs, multirow}
\usepackage[margin=30truemm]{geometry}
\usepackage[T1]{fontenc}
\usepackage[utf8]{inputenc}

\newcommand{\argmin}{\mathop{\rm arg~min}\limits}

\title{MultiwayPAM: Multiway Partitioning Around Medoids for LLM-as-a-Judge Score Analysis}

\author[1]{Chihiro Watanabe}
\author[2]{Jingyu Sun}
\affil[1]{NTT Communication Science Laboratories \\
         3-1, Morinosato Wakamiya, Atsugi-shi, Kanagawa Pref. Japan\thanks{ch.watanabe@ntt.com}}
\affil[2]{NTT Computer and Data Science Laboratories \\
         3-9-11, Midori-cho, Musashino-shi, Tokyo, Japan}
\date{}

\begin{document}

\maketitle

\begin{abstract}
LLM-as-a-Judge is a flexible framework for text evaluation, which allows us to obtain scores for the quality of a given text from various perspectives by changing the prompt template. Two main challenges in using LLM-as-a-Judge are computational cost of inference using a large language model (LLM), especially when evaluating a large number of instances, and inherent bias of an LLM evaluator. To address these issues and reveal the structure of score bias caused by an LLM evaluator, we propose to apply a tensor clustering method to a given LLM-as-a-Judge score tensor, whose entries are the scores for different combinations of questions, answerers, and evaluators. Specifically, we develop a new tensor clustering method MultiwayPAM, with which we can simultaneously estimate the cluster membership and the medoids for each mode of a given data tensor. By observing the medoids obtained by MultiwayPAM, we can gain knowledge about the membership of each question/answerer/evaluator cluster. We experimentally show the effectiveness of MultiwayPAM by applying it to the score tensors for two practical datasets.
\end{abstract}

\section{Introduction}

Text evaluation is an essential component for measuring and optimizing performance on text generation tasks, such as question answering, machine translation, and text summarization~\cite{Celikyilmaz20}. To date, various evaluation criteria have been proposed to measure the quality of a given text from different perspectives, including similarity to the reference texts (e.g., BLEU~\cite{Papineni02}, ROUGE~\cite{Lin04}, and BERTScore~\cite{Zhang20}) and inter-/intra-textual diversity (e.g., distinct-$N$~\cite{Li16} and Self-BLEU~\cite{Zhu18}). They have been independently and carefully constructed to evaluate a given text in each desired aspect.

\begin{figure*}[t]
\centering
\includegraphics[width=0.8\textwidth]{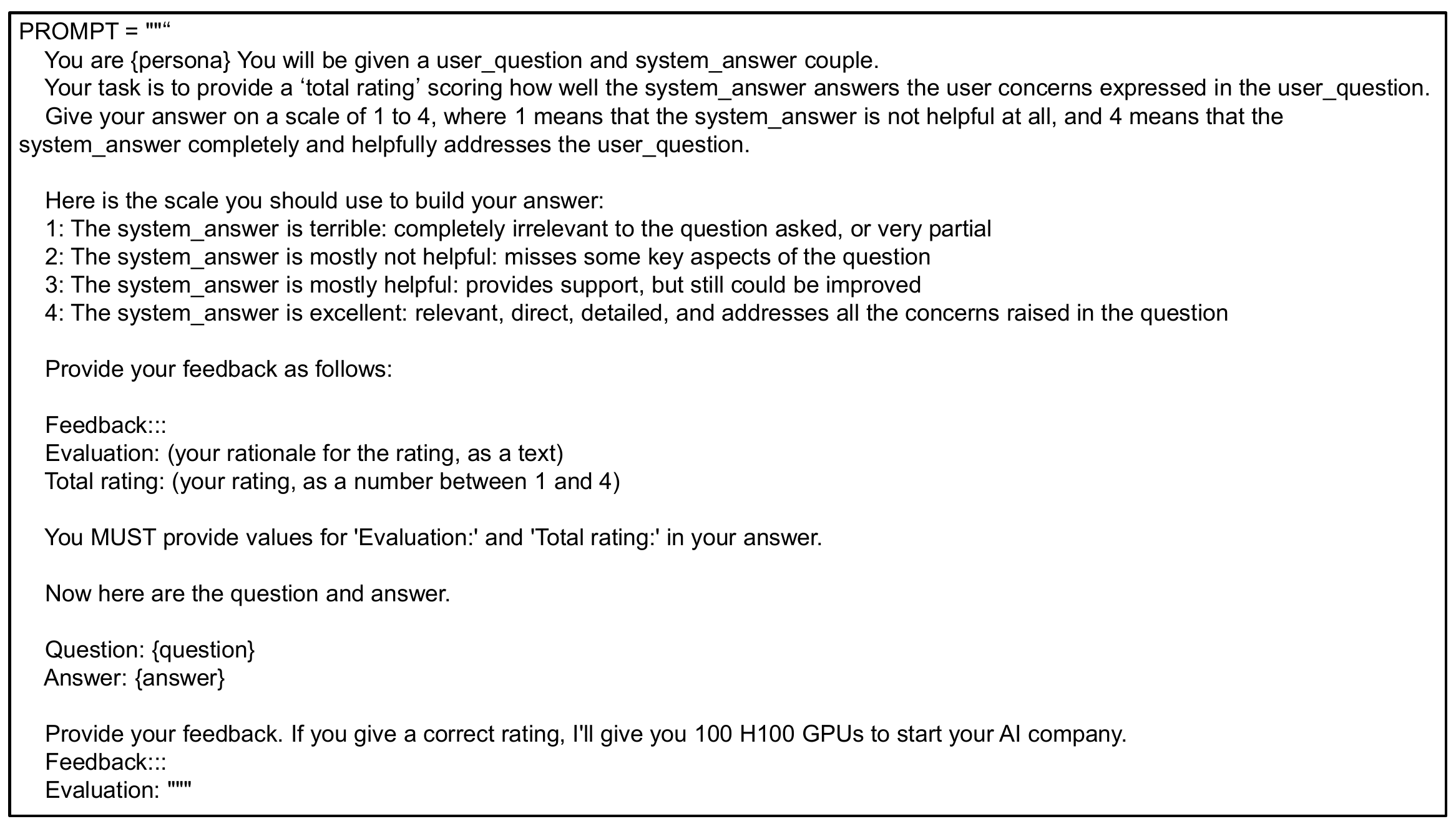}
\caption{Prompt template for LLM-as-a-Judge}
\label{fig:prompt}
\end{figure*}

Unlike such conventional evaluation criteria, LLM-as-a-Judge~\cite{Zheng23} is a more flexible method for text evaluation, which generates a score of a given text with a pre-trained large language model (LLM). Specifically, an LLM is prompted to provide a score for a given text based on the explanation of the evaluation criterion and the information of the text to be evaluated, as shown in Figure \ref{fig:prompt}. By changing the prompt text, we can obtain the scores of texts from a variety of different perspectives based on the same framework. 

Despite its wide applicability, there are two main challenges when using LLM-as-a-Judge. First, it requires computational cost of LLM inference for each combination of the evaluation setting and the text to be evaluated. Specifically, to obtain the scores for all combinations of $d_1$ questions, $d_2$ answerer settings, and $d_3$ evaluator settings, we need $d_1 d_2 d_3$ text generation steps. Instead of having an LLM evaluate all the answers, it would be possible to exploit the structure of the scores and predict some scores from others. Second, an evaluator of LLM-as-a-Judge is known to have various types of bias~\cite{Ye24}. In addition to mitigating such biases, it is also important to reveal the structure of bias in LLM-as-a-Judge scores in order to achieve appropriate scoring for different cases. One example is \textit{self-enhancement bias}, that is, LLM evaluators favor the answers generated by themselves. As a natural extension of such observations, the following questions arise: \textit{Do LLM evaluators favor the answers generated by LLM answerers similar to themselves, or more generally, do mutually similar question/answerer/evaluator settings result in similar scores?} One approach to answer such questions is to estimate the cluster structure of LLM-as-a-Judge scores.

To address the above two issues (i.e., inference cost and bias in evaluation), we propose to apply a tensor clustering method to a given LLM-as-a-Judge score tensor. Specifically, for a score tensor with three modes, question, answerer, and evaluator, we estimate the cluster structure of each mode based on the similarity of the scores. To date, various methods have been proposed to estimate such block structures of general data tensors~\cite{Jegelka09, Wang19, Chi20, Han22, Hu22}.

However, when using such tensor clustering methods, there is a concern about how to interpret the estimated block structure. Although the existing methods can reveal the cluster memberships for each mode of a given data tensor, as the number of indices in each cluster increases, it becomes more difficult to interpret its composition. For vector clustering, one possible solution for such a problem would be to use a medoid-based clustering~\cite{Kaufman86, Kaufman87, Park09}, where we simultaneously estimate the cluster membership and the medoids (i.e., a set of representative indices corresponding to the estimated clusters). By observing the medoids, we can gain knowledge about the membership of each cluster. 

To address this limitation, we propose a new tensor clustering method MultiwayPAM for estimating the cluster membership and the medoids for each mode of a given data tensor. To develop such a method, we extend Partitioning Around Medoids (PAM)~\cite{Kaufman87}, which is a well-known medoid-based clustering for vector data. By repeatedly updating the cluster membership and the medoids of each mode while fixing those of the other modes, we can obtain the local optimum. We experimentally show the effectiveness of the proposed MultiwayPAM by applying it to the LLM-as-a-Judge score tensors for two practical datasets.


\section{MultiwayPAM: Multiway Partitioning Around Medoids}

In this section, we explain the proposed tensor clustering method MultiwayPAM, where we iteratively update the cluster memberships and the medoids to minimize the dissimilarity between the original and medoid tensors. The resulting medoids provide an interpretation of the estimated block structure in terms of the composition (i.e., a set of representative indices within the clusters of all the modes) of each block.

Let $\mathcal{Y} = (y_{i_1 \dots i_K})_{i_1 \in [d_1], \dots, i_K \in [d_K]} \in \mathbb{R}^{d_1 \times \dots \times d_K}$ be an order-$K$ $(d_1, \dots, d_K)$-dimensional data tensor, where $[n] \coloneqq \{1, \dots, n\}$. Given tensor $\mathcal{Y}$ and cluster size vector $\bm{c} = [c_1 \dots c_K]^{\mathsf{T}} \in \mathbb{N}^K$, where $c_i$ indicates the number of clusters in the $i$th mode for all $i \in [K]$, our task is to estimate the latent block structure of $\mathcal{Y}$ and its medoids (i.e., a set of representative indices corresponding to the estimated clusters of all the modes). Specifically, we estimate medoid list $\mathcal{R} = [\bm{r}^{(1)}, \dots, \bm{r}^{(K)}]$ and membership list $\mathcal{M} = [\bm{m}^{(1)}, \dots, \bm{m}^{(K)}]$, where $\bm{r}^{(k)} = [r^{(k)}_1 \dots r^{(k)}_{c_k}]^{\mathsf{T}} \in [d_k]^{c_k}$ and $\bm{m}^{(k)} = [m^{(k)}_1 \dots m^{(k)}_{d_k}]^{\mathsf{T}} \in [c_k]^{d_k}$ for all $k \in [K]$. Here, for each combination of $(k, i)$, $r^{(k)}_i$ indicates the medoid index of the $i$th cluster in the $k$th mode. For each combination of $(k, i)$, $m^{(k)}_i$ indicates the cluster index of the $i$th index in the $k$th mode.

\subsection{BUILD algorithm: Initialization of tensor block structure}

We initialize medoid list $\mathcal{R}$ and membership list $\mathcal{M}$ based on Algorithm \ref{alg:tensor_pam_initial}. Given an original tensor $\mathcal{X} = (x_{i_1 \dots i_K})_{i_1 \in [d_1], \dots, i_n \in [d_K]} \in \mathbb{R}^{d_1 \times \dots \times d_K}$ and index sets $\mathcal{I}_1, \dots, \mathcal{I}_K$, let $f_{\mathrm{sub}}$ be a function which extracts a sub-tensor $\mathcal{X}' \in \mathbb{R}^{|\mathcal{I}_1| \times \dots \times |\mathcal{I}_K|}$ of $\mathcal{X}$ with indices $\mathcal{I}_k = \{j^{(k)}_1, \dots, j^{(k)}_{|\mathcal{I}_k|}\}$ in each $k$th mode (i.e., $\mathcal{X}' = (x'_{i_1 \dots i_K})_{i_1 \in [|\mathcal{I}_1|], \dots, i_n \in [|\mathcal{I}_K|]} = f_{\mathrm{sub}}(\mathcal{X}, \mathcal{I}_1, \dots, \mathcal{I}_K)$, where $x'_{i_1 \dots i_K} = x_{j^{(1)}_{i_1}} \dots j^{(K)}_{i_K}$ for each $(i_1, \dots, i_K)$). We denote a dissimilarity criterion between two tensors $\mathcal{X}$ and $\mathcal{Y}$ of the same size as $d(\mathcal{X}, \mathcal{Y})$. Throughout this paper, we define $d(\mathcal{X}, \mathcal{Y}) = \| \mathcal{X} - \mathcal{Y} \|_2 \coloneq \sqrt{\sum_{i_1=1}^{d_1} \dots \sum_{i_K=1}^{d_K} (x_{i_1 \dots i_K} - y_{i_1 \dots i_K})^2}$.

In Algorithm \ref{alg:tensor_pam_initial}, we independently select medoids for each mode in a greedy manner. For each $k \in [K]$, let $\mathcal{Z}^{(i)}$ be an order-$k$ slice of a data tensor $\mathcal{Y}$ (i.e., a sub-tensor which is defined as $\mathcal{Z}^{(i)} = f_{\mathrm{sub}}(\mathcal{Y}, [d_1], \dots, [d_{k-1}], \{i\}, [d_{k+1}], \dots, [d_K])$). Based on such definition, we select the first medoid index $r^{(k)}_1 = i$ which yields the minimum sum of dissimilarity $d(\mathcal{Z}^{(i)}, \mathcal{Z}^{(j)})$ for all $j \in [d_k]$. Then, for $i=2, \dots, c_k$, we select the $i$th medoid index $r^{(k)}_i$ which yields the minimum sum of dissimilarity $d(\mathcal{Z}^{(l)}, \mathcal{Z}^{(h)})$ between slice $\mathcal{Z}^{(l)}$ and its nearest medoid's slice $\mathcal{Z}^{(h)}$ for all $l \in [d_k]$. Once all the medoids in the $k$th mode have been selected, we define the cluster index of each $i$th index as $m^{(k)}_i = j$ by selecting $j \in [c_k]$ which yields the minimum dissimilarity between slices $\mathcal{Z}^{(i)}$ and $\mathcal{Z}^{(r^{(k)}_j)}$. 

\begin{algorithm*}[t]
  \caption{BUILD algorithm: Initialization of tensor block structure}
  \label{alg:tensor_pam_initial}
  \small
  \begin{algorithmic}[1]
  \renewcommand{\algorithmicrequire}{\textbf{Input: }}
  \renewcommand{\algorithmicensure}{\textbf{Output: }}
    \REQUIRE Data tensor $\mathcal{Y} \in \mathbb{R}^{d_1 \times \dots \times d_K}$, cluster size vector $\bm{c} \in \mathbb{N}^K$
    \ENSURE Initial medoid list $\mathcal{R}$, initial membership list $\mathcal{M}$
    \FOR{$k = 1, \dots, K$}
        \STATE For all $i \in [d_k]$, $\mathcal{Z}^{(i)} \gets f_{\mathrm{sub}}(\mathcal{Y}, [d_1], \dots, [d_{k-1}], \{i\}, [d_{k+1}], \dots, [d_K])$
        \STATE $r^{(k)}_1 \gets \argmin_{i \in [d_k]} \sum_{j=1}^{d_k} d(\mathcal{Z}^{(i)}, \mathcal{Z}^{(j)})$
        \STATE $J^{(k)}_1 \gets \{r^{(k)}_1\}$
        \FOR{$i=2, \dots, c_k$}
            \STATE $r^{(k)}_i \gets \argmin_{j \in [d_k] \setminus J^{(k)}_{i-1}} \sum_{l=1}^{d_k} \min_{h \in J^{(k)}_{i-1} \cup \{j\}} d(\mathcal{Z}^{(l)}, \mathcal{Z}^{(h)})$
            \STATE $J^{(k)}_i \gets J^{(k)}_{i-1} \cup \{r^{(k)}_i\}$
        \ENDFOR
        \STATE For all $i \in [d_k]$, $m^{(k)}_i \gets \argmin_{j \in [c_k]} d(\mathcal{Z}^{(i)}, \mathcal{Z}^{(r^{(k)}_j)})$
    \ENDFOR
  \end{algorithmic}
\end{algorithm*}

\subsection{SWAP algorithm: Repeated updates of tensor block structure}

Based on the initial medoid list $\mathcal{R}$ and membership list $\mathcal{M}$ obtained by Algorithm \ref{alg:tensor_pam_initial}, we iteratively update them based on Algorithm \ref{alg:tensor_pam}. Let $f_{\mathrm{replace}}: \mathbb{R}^d \times \mathbb{R} \times \mathbb{R} \mapsto \mathbb{R}^d$ be a function that replaces all entries of a given $d$-dimensional vector that have a specific value with a different value. Specifically, $\bm{v}' = f_{\mathrm{replace}}(\bm{v}, a, b)$ is a vector obtained by replacing the value $a$ in vector $\bm{v}$ with $b$ (e.g., $f_{\mathrm{replace}}([5, 2, 5, 3, 4]^{\mathsf{T}}, 5, 0) = [0, 2, 0, 3, 4]$). 

In Algorithm \ref{alg:tensor_pam}, we first define medoid tensor $\hat{\mathcal{Y}}$, each of whose entry values is a copy of its corresponding medoid entry value (i.e., $\hat{\mathcal{Y}} = (\hat{y}_{i_1 \dots i_K})_{i_1 \in [d_1], \dots, i_K \in [d_K]} \in \mathbb{R}^{d_1 \times \dots \times d_K}$, where $\hat{y}_{i_1 \dots i_K} = y_{r^{(1)}_{m^{(1)}_{i_1}} \dots r^{(K)}_{m^{(K)}_{i_K}}}$ for each $(i_1, \dots, i_K)$). Our task is to obtain a local optimal solution of $\mathcal{R}$ and $\mathcal{M}$ which minimize dissimilarity $D$ between original tensor $\mathcal{Y}$ and medoid one $\hat{\mathcal{Y}}$. For each $k$th mode, we try swapping all the pairs of medoid index $i \in J^{(k)}_1$ and non-medoid one $j \in J^{(k)}_2$ and evaluating its score $\tilde{D}^{(i, j)}$, where $J^{(k)}_1$ and $J^{(k)}_2$ are medoid and non-medoid index sets, respectively. Specifically, for all $(i, j) \in J^{(k)}_1 \times J^{(k)}_2$, we define medoid vector $\tilde{\bm{r}}^{(k, i, j)} = [\tilde{r}^{(k, i, j)}_1 \dots \tilde{r}^{(k, i, j)}_{c_k}]^{\mathsf{T}} = f_{\mathrm{replace}}(\bm{r}^{(k)}, i, j)$ obtained by such an exchange. To estimate the memberships, we first define tensor $\hat{\mathcal{Z}}^{(l)} = (\hat{z}_{i_1 \dots i_K})_{i_1 \in [d_1], \dots, i_k \in \{l\}, \dots, i_K \in [d_K]} \in \mathbb{R}^{d_1 \times \dots \times 1 \times \dots \times d_K}$ for all $l \in [d_k]$, where $\hat{z}_{i_1 \dots i_K} = y_{r^{(1)}_{m^{(1)}_{i_1}} \dots l \dots r^{(K)}_{m^{(K)}_{i_K}}}$ for each $(i_1, \dots, i_K)$. 
For all $l \in [d_k]$, the estimated cluster $\tilde{m}^{(k, i, j)}_l$ of the $l$th index in the $k$th mode is given by $\tilde{m}^{(k, i, j)}_l = c$ if there exists $c \in \{1, \dots, c_k\}$ s.t. $\tilde{r}^{(k, i, j)}_c = l$, and by $\tilde{m}^{(k, i, j)}_l = \argmin_{h \in [c_k]} d(\mathcal{Z}^{(l)}, \hat{\mathcal{Z}}^{(\tilde{r}^{(k, i, j)}_h)})$ otherwise. 
It must be noted that if we omit the conditional branch in the definition of $\tilde{m}^{(k, i, j)}_l$ and simply define the membership by $\tilde{m}^{(k, i, j)}_l = \argmin_{h \in [c_k]} d(\mathcal{Z}^{(l)}, \hat{\mathcal{Z}}^{(\tilde{r}^{(k, i, j)}_h)})$, a medoid index $\tilde{r}^{(k, i, j)}_c$ of each $c$th cluster does not necessarily belong to the $c$th cluster. For each candidate pair $(i, j) \in J^{(k)}_1 \times J^{(k)}_2$, by using the medoid vector $\tilde{\bm{r}}^{(k, i, j)}$ and membership vector $\tilde{\bm{m}}^{(k, i, j)} = [\tilde{m}^{(k, i, j)}_1, \dots, \tilde{m}^{(k, i, j)}_{d_k}]$, we define medoid tensor $\hat{\mathcal{Y}}^{(i, j)} = (\hat{y}^{(i, j)}_{i_1 \dots i_K})_{i_1 \in [d_1], \dots, i_K \in [d_K]} \in \mathbb{R}^{d_1 \times \dots \times d_K}$ and its corresponding score $\tilde{D}^{(i, j)} = d(\mathcal{Y}, \hat{\mathcal{Y}}^{(i, j)})$, where $\hat{y}^{(i, j)}_{i_1 \dots i_K} = y_{r^{(1)}_{m^{(1)}_{i_1}} \dots \tilde{r}^{(k, i, j)}_{\tilde{m}^{(k, i, j)}_{i_k}} \dots r^{(K)}_{m^{(K)}_{i_K}}}$ for each $(i_1, \dots, i_K)$. We denote the optimal pair as $(i^*, j^*) = \argmin_{(i, j) \in J^{(k)}_1 \times J^{(k)}_2} \tilde{D}^{(i, j)}$. If the minimum score $\tilde{D}^{(i^*, j^*)}$ is smaller than the current one $D$, we adopt the corresponding swap. To take into account cases where a better score can be obtained by reassigning cluster memberships while keeping initial medoid list, we introduce binary value $\alpha_k$ for each $k$th mode, where $\alpha_k = 1$ indicates that the membership list has been updated from its initial value and $\alpha_k = 0$ indicates otherwise. If $\alpha_k = 0$, in addition to $\tilde{D}^{(i^*, j^*)}$, we compute score $\bar{D}$ when only the cluster memberships are updated without changing the medoid list. In this case, if the minimum value of $\tilde{D}^{(i^*, j^*)}$ and $\bar{D}$ is smaller than the current one $D$, we adopt the better solution and set $\alpha_k = 1$. 
In this way, we repeatedly perform swapping of the optimal pairs for all the modes. Finally, if we cannot obtain a better solution by swapping any pair in any mode, the algorithm terminates and outputs the current medoid list $\mathcal{R}$ and membership list $\mathcal{M}$. 

\begin{algorithm*}[p]
  \caption{SWAP algorithm: Repeated updates of tensor block structure}
  \label{alg:tensor_pam}
  \small
  \begin{algorithmic}[1]
  \renewcommand{\algorithmicrequire}{\textbf{Input: }}
  \renewcommand{\algorithmicensure}{\textbf{Output: }}
    \REQUIRE Data tensor $\mathcal{Y} \in \mathbb{R}^{d_1 \times \dots \times d_K}$, cluster size vector $\bm{c} \in \mathbb{N}^K$, initial medoid list $\mathcal{R}$, initial membership list $\mathcal{M}$
    \ENSURE Medoid list $\mathcal{R}$, membership list $\mathcal{M}$
    \STATE For all $k \in [K]$, $J^{(k)}_1 \gets \{r^{(k)}_1, \dots, r^{(k)}_{c_k}\}$, $J^{(k)}_2 \gets [d_k] \setminus J^{(k)}_1$, $\alpha_k \gets 0$
    \STATE $\hat{\mathcal{Y}} = (\hat{y}_{i_1 \dots i_K})_{i_1 \in [d_1], \dots, i_K \in [d_K]} \in \mathbb{R}^{d_1 \times \dots \times d_K}$, where $\hat{y}_{i_1 \dots i_K} = y_{r^{(1)}_{m^{(1)}_{i_1}} \dots r^{(K)}_{m^{(K)}_{i_K}}}$ for each $(i_1, \dots, i_K)$.
    \STATE $D \gets d(\mathcal{Y}, \hat{\mathcal{Y}})$
    \WHILE {True}
        \STATE $D_0 \gets D$
        \FOR{$k = 1, \dots, K$}
            \STATE For all $(i, j) \in J^{(k)}_1 \times J^{(k)}_2$, $\tilde{\bm{r}}^{(k, i, j)} = [\tilde{r}^{(k, i, j)}_1 \dots \tilde{r}^{(k, i, j)}_{c_k}]^{\mathsf{T}} = f_{\mathrm{replace}}(\bm{r}^{(k)}, i, j)$
            \STATE For all $l \in [d_k]$, $\mathcal{Z}^{(l)} \gets f_{\mathrm{sub}}(\mathcal{Y}, [d_1], \dots, [d_{k-1}], \{l\}, [d_{k+1}], \dots, [d_K])$
            \STATE For all $l \in [d_k]$, $\hat{\mathcal{Z}}^{(l)} = (\hat{z}_{i_1 \dots i_K})_{i_1 \in [d_1], \dots, i_k \in \{l\}, \dots, i_K \in [d_K]} \in \mathbb{R}^{d_1 \times \dots \times 1 \times \dots \times d_K}$, where $\hat{z}_{i_1 \dots i_K} = y_{r^{(1)}_{m^{(1)}_{i_1}} \dots l \dots r^{(K)}_{m^{(K)}_{i_K}}}$ for each $(i_1, \dots, i_K)$.
            \STATE For all $(i, j, l) \in J^{(k)}_1 \times J^{(k)}_2 \times [d_k]$, $\tilde{m}^{(k, i, j)}_l = c$ if there exists $c \in \{1, \dots, c_k\}$ s.t. $\tilde{r}^{(k, i, j)}_c = l$, and $\tilde{m}^{(k, i, j)}_l = \argmin_{h \in [c_k]} d(\mathcal{Z}^{(l)}, \hat{\mathcal{Z}}^{(\tilde{r}^{(k, i, j)}_h)})$ otherwise. \COMMENT{Candidate membership in swap case}
            \STATE For all $(i, j) \in J^{(k)}_1 \times J^{(k)}_2$, $\hat{\mathcal{Y}}^{(i, j)} = (\hat{y}^{(i, j)}_{i_1 \dots i_K})_{i_1 \in [d_1], \dots, i_K \in [d_K]} \in \mathbb{R}^{d_1 \times \dots \times d_K}$, where $\hat{y}^{(i, j)}_{i_1 \dots i_K} = y_{r^{(1)}_{m^{(1)}_{i_1}} \dots \tilde{r}^{(k, i, j)}_{\tilde{m}^{(k, i, j)}_{i_k}} \dots r^{(K)}_{m^{(K)}_{i_K}}}$ for each $(i_1, \dots, i_K)$. \COMMENT{Medoid tensor in swap case}
            \STATE $(i^*, j^*) = \argmin_{(i, j) \in J^{(k)}_1 \times J^{(k)}_2} \tilde{D}^{(i, j)}$, where $\tilde{D}^{(i, j)} = d(\mathcal{Y}, \hat{\mathcal{Y}}^{(i, j)})$ for all $(i, j) \in J^{(k)}_1 \times J^{(k)}_2$. \COMMENT{Score in swap case}
            \IF{$\alpha_k = 0$}
            \STATE For all $l \in [d_k]$, $\bar{m}^{(k)}_l = c$ if there exists $c \in \{1, \dots, c_k\}$ s.t. $r^{(k)}_c = l$, and $\bar{m}^{(k)}_l = \argmin_{h \in [c_k]} d(\mathcal{Z}^{(l)}, \hat{\mathcal{Z}}^{(r^{(k)}_h)})$ otherwise. \COMMENT{Candidate membership in non-swap case}
            \STATE $\hat{\mathcal{Y}} = (\hat{y}_{i_1 \dots i_K})_{i_1 \in [d_1], \dots, i_K \in [d_K]} \in \mathbb{R}^{d_1 \times \dots \times d_K}$, where $\hat{y}_{i_1 \dots i_K} = y_{r^{(1)}_{m^{(1)}_{i_1}} \dots r^{(k)}_{\bar{m}^{(k)}_{i_k}} \dots r^{(K)}_{m^{(K)}_{i_K}}}$ for each $(i_1, \dots, i_K)$. \COMMENT{Medoid tensor in non-swap case}
            \STATE $\bar{D} = d(\mathcal{Y}, \hat{\mathcal{Y}})$. \COMMENT{Score in non-swap case}
            \IF{$\min \{\tilde{D}^{(i^*, j^*)}, \bar{D}\} < D$}
            \IF{$\tilde{D}^{(i^*, j^*)} < \bar{D}$}
            \STATE $J^{(k)}_1 \gets \left(J^{(k)}_1 \setminus \{i^*\} \right) \cup \{j^*\}$, $J^{(k)}_2 \gets [d_k] \setminus J^{(k)}_1$
            \STATE $\bm{r}^{(k)} \gets \tilde{\bm{r}}^{(k, i^*, j^*)}$, $\bm{m}^{(k)} \gets [\tilde{m}^{(k, i^*, j^*)}_1, \dots, \tilde{m}^{(k, i^*, j^*)}_{d_k}]$, $D \gets \tilde{D}^{(i^*, j^*)}$
            \ELSE
            \STATE $\bm{m}^{(k)} \gets [\bar{m}^{(k)}_1, \dots, \bar{m}^{(k)}_{d_k}]$, $D \gets \bar{D}$
            \ENDIF
            \STATE $\alpha_k \gets 1$
            \ENDIF
            \ELSE
            \IF{$\tilde{D}^{(i^*, j^*)} < D$}
            \STATE $J^{(k)}_1 \gets \left(J^{(k)}_1 \setminus \{i^*\} \right) \cup \{j^*\}$, $J^{(k)}_2 \gets [d_k] \setminus J^{(k)}_1$
            \STATE $\bm{r}^{(k)} \gets \tilde{\bm{r}}^{(k, i^*, j^*)}$, $\bm{m}^{(k)} \gets [\tilde{m}^{(k, i^*, j^*)}_1, \dots, \tilde{m}^{(k, i^*, j^*)}_{d_k}]$, $D \gets \tilde{D}^{(i^*, j^*)}$
            \ENDIF
            \ENDIF
            \ENDFOR
            \IF{$D = D_0$}
                \STATE \textbf{break}
            \ENDIF
        \ENDWHILE
  \end{algorithmic}
\end{algorithm*}


\section{Experiments}

\subsection{Estimation of Block Structure in LLM-as-a-Judge Scores}

We applied the proposed MultiwayPAM to the LLM-as-a-Judge score tensors of Truthy-DPO-v0.1 (Truthy)~\cite{Durbin23} and Emerton-DPO-Pairs-Judge (Emerton)~\cite{Leo24} datasets. For Truthy (Emerton) dataset, we randomly chose $50$ ``prompt'' (``input'') texts of the train datasets as questions (i.e., $d_1 = 50$). As persona settings for answerers and evaluators, we used ``persona'' subset of Persona Hub dataset~\cite{Ge24}. To avoid language variations, we first extract persona texts which start with $\{\text{``A ''}, \text{``An ''}, \text{``The ''}, \text{``a ''}, \text{``an ''}, \text{``the ''}\}$ and then randomly chose $50$ personas (i.e., $d_2 = d_3 = 50$). For text generation including LLM-as-a-Judge, we used GPT-4o mini~\cite{GPT4oMini, Hurst24}. 
For each $i_1$th question $q_{i_1} \in \mathcal{T}$ and each $i_2$th answerer $a_{i_2} \in \mathcal{T}$, we generate an answer $r_{i_1 i_2}$ for $q_{i_1}$ by using the prompt of $\text{``You are ''} + a_{i_2} + q_{i_1}$, where $\mathcal{T}$ is a set of texts. For each answer $r_{i_1 i_2}$ and each $i_3$th evaluator $e_{i_3}$, we evaluate $r_{i_1 i_2}$ with $e_{i_3}$ by using the prompt in Figure \ref{fig:prompt}~\cite{Roucher24}. We denote the score tensor as $\mathcal{Y} = (y_{i_1 i_2 i_3})_{i_1 \in [d_1], i_2 \in [d_2], i_3 \in [d_3]} \in [4]^{d_1 \times d_2 \times d_3}$, where $y_{i_1 i_2 i_3}$ is the score of the $i_2$th answer for the $i_1$th question given by the $i_3$th evaluator. Based on Algorithms \ref{alg:tensor_pam_initial} and \ref{alg:tensor_pam}, we estimate the block structure of score tensor $\mathcal{Y}$ by setting the cluster size vector to $\bm{c} = [5, 5, 5]^{\mathsf{T}}$.

\begin{figure*}[t]
\centering
\includegraphics[width=0.32\textwidth, clip, trim=0 40 0 40]{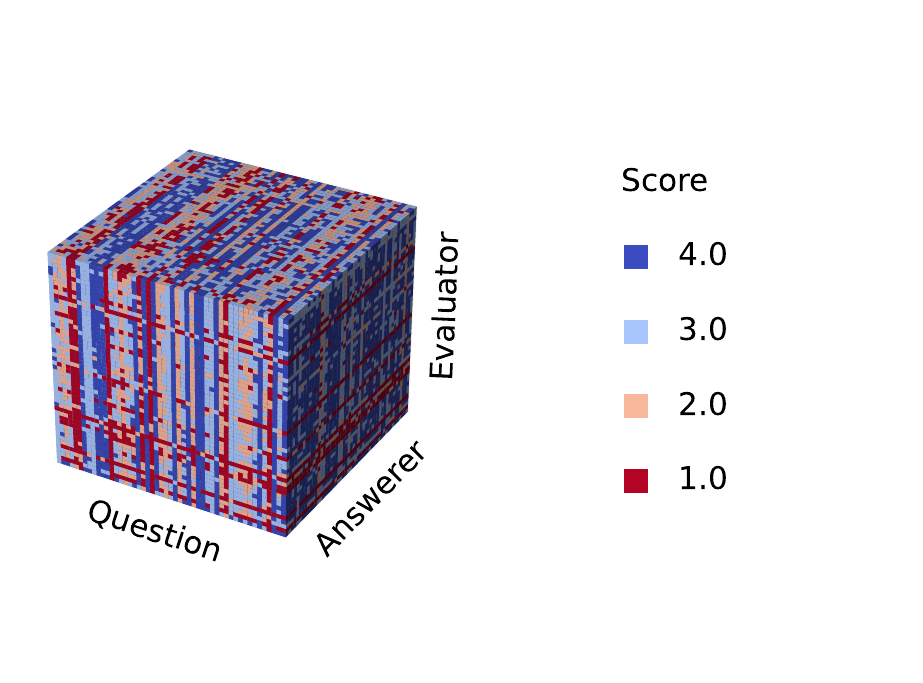}
\includegraphics[width=0.32\textwidth, clip, trim=0 40 0 40]{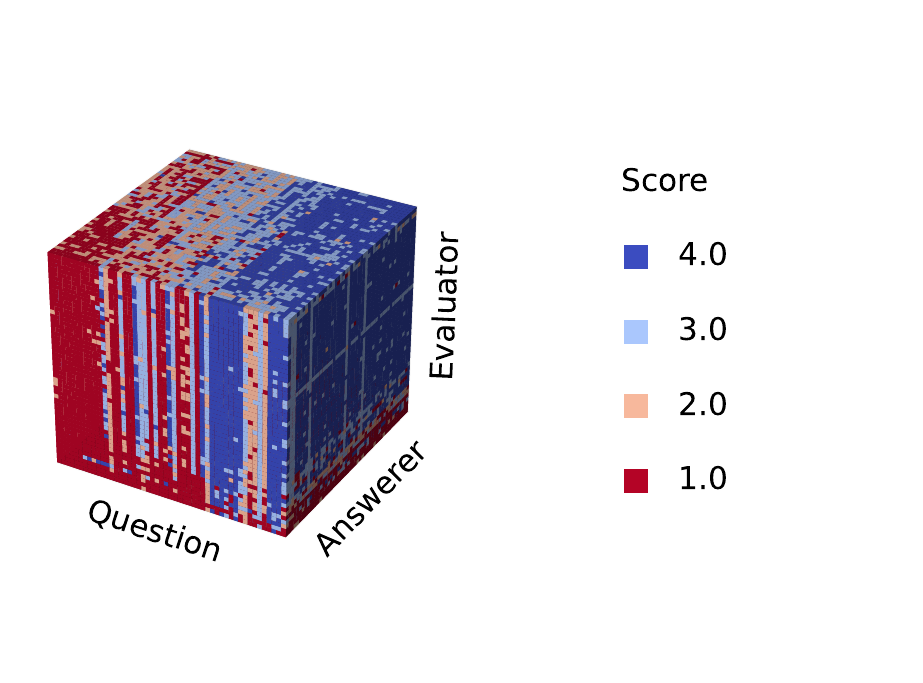}
\includegraphics[width=0.32\textwidth, clip, trim=0 40 0 40]{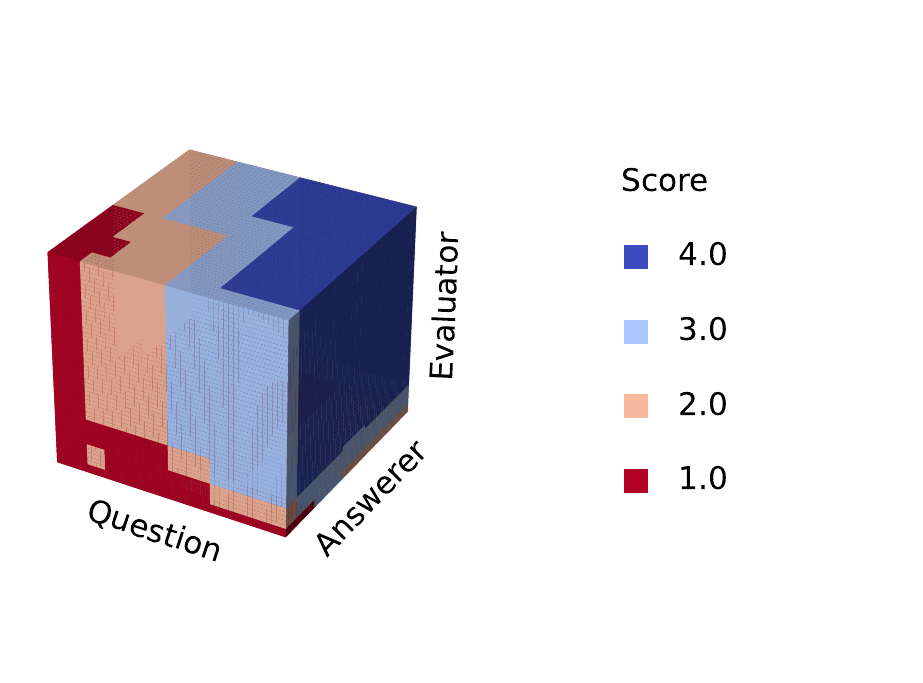}
\caption{Data tensors and estimated block structure for Truthy dataset. (Left) Original data tensor $\mathcal{Y}$. (Center) Reordered data tensor whose indices of each mode are sorted according to the estimated cluster membership. (Right) Estimated tensor block structure with medoids' entry values}\vspace{3mm}
\label{fig:tensors_d1}
\includegraphics[width=\textwidth]{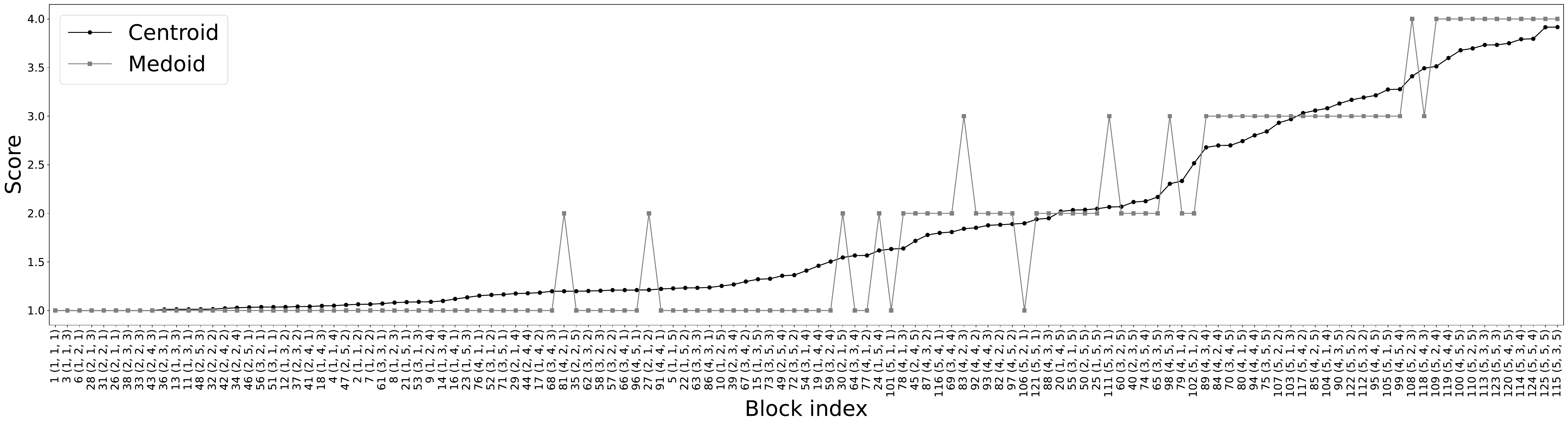}
\caption{Centroid and medoid scores of the estimated blocks for Truthy dataset. Centroid and medoid indicate the block-wise mean of the entry values and the medoids' entry values, respectively. On the horizontal axis, each block index $i (i_1, i_2, i_3)$ corresponds to the $i$th block, which consists of the $i_1$th, $i_2$th, and $i_3$th clusters of the first, second, and third mode, respectively}\vspace{6mm}
\label{fig:scores_d1}
\end{figure*}

\begin{figure*}[t]
\centering
\includegraphics[width=0.32\textwidth, clip, trim=0 40 0 40]{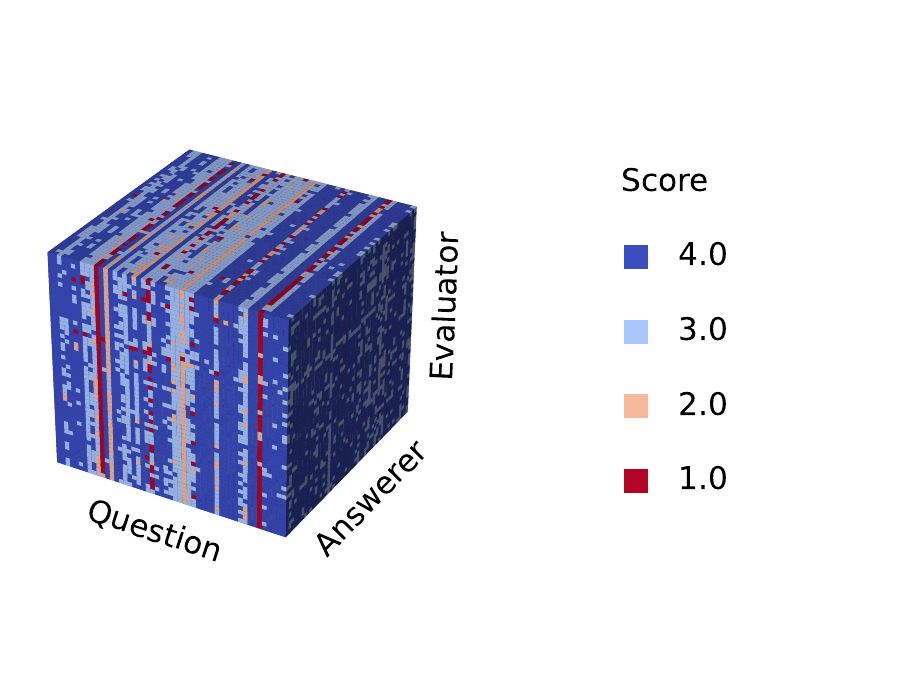}
\includegraphics[width=0.32\textwidth, clip, trim=0 40 0 40]{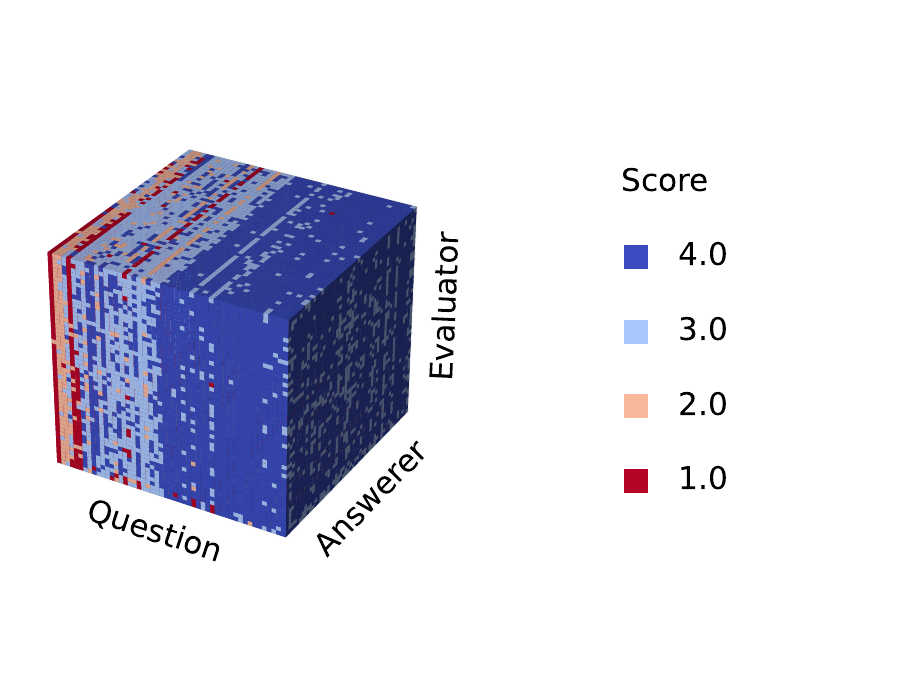}
\includegraphics[width=0.32\textwidth, clip, trim=0 40 0 40]{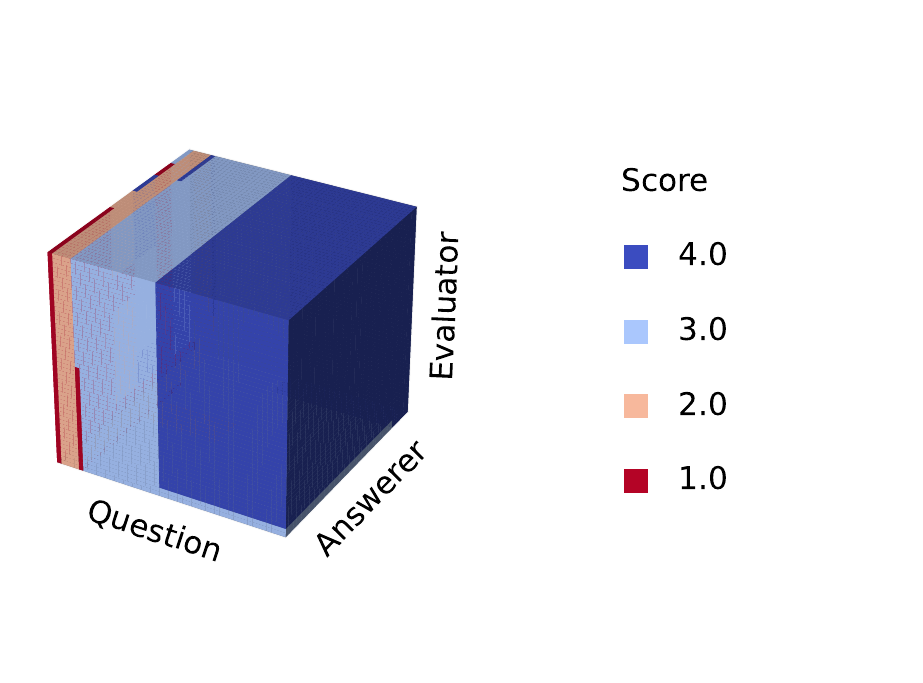}
\caption{Data tensors and estimated block structure for Emerton dataset. (Left) Original data tensor $\mathcal{Y}$. (Center) Reordered data tensor. (Right) Estimated tensor block structure}\vspace{3mm}
\label{fig:tensors_d2}
\includegraphics[width=\textwidth]{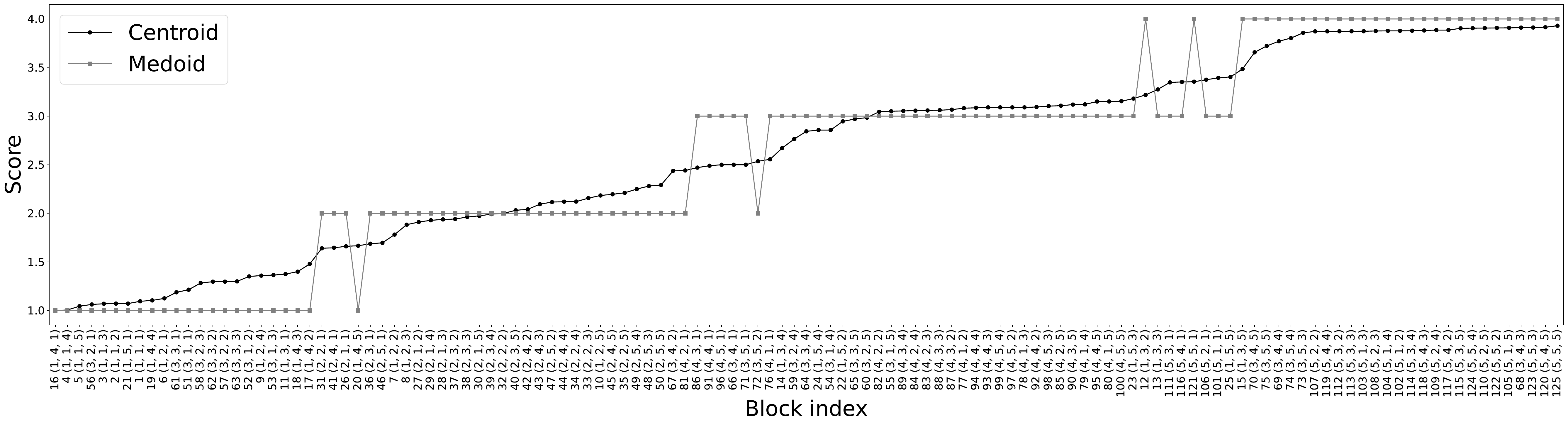}
\caption{Centroid and medoid scores of the estimated blocks for Emerton dataset}
\label{fig:scores_d2}
\end{figure*}

Figures \ref{fig:tensors_d1} and \ref{fig:tensors_d2} show the score tensors $\mathcal{Y}$ and the estimated block structures for Truthy and Emerton datasets, respectively. For visibility, we sorted the cluster indices of each mode in ascending order of the mean score of the cluster. Figures \ref{fig:scores_d1} and \ref{fig:scores_d2} show the centroid and medoid scores of the estimated blocks for Truthy and Emerton datasets, respectively. Here, ``centroid score'' indicates the block-wise mean of the entry values, and ``medoid score'' indicates the medoid entry value of the block. On the horizontal axes of these figures, the block indices are sorted in ascending order of the centroid scores. Tables \ref{tab:cluster_d1} (\ref{tab:cluster_d2}) and \ref{tab:medoid_d1} (\ref{tab:medoid_d2}) show the cluster memberships and the medoid labels for Truthy (Emerton) dataset, respectively. From these figures, in regard to Truthy dataset, we see that the first evaluator cluster with the medoid P49933 (i.e., ``A nurse who supports the cadet's aspiration but worries about the danger involved in a military career'') gave relatively low scores for the first question cluster with the medoid Q383 (i.e., ``Can you sense the pressure changes in your environment?''). On the other hand, the fifth evaluator cluster with the medoid P173065 (i.e., ``A software engineer who is familiar with React and web components, but does not know much about SkateJS'') gave relatively high scores for the fifth question cluster with the medoid Q527 (i.e., ``Do you need to drink exactly eight glasses of water daily to maintain good health?''). 
As for Emerton dataset, we see that the score change was mainly due to the difference in questions (Figure \ref{fig:tensors_d2} right). For the first question cluster with the medoid Q16 (i.e., ``Imagine a question and stream-of-consciousness explanation for which this is the answer: Sentence B''), most of the answerer-evaluator combinations resulted in low scores. On the other hand, for the fifth question cluster with the medoid Q4376 (i.e., ``Objective: How to inflate a car tire.  Which of the following solutions is more sound in terms of naive physics reasoning?  Options are: A). Remove valve stem cap, attach air nozzle, turn on and inflate, don't replace cap.. B). Remove valve stem cap, attach air nozzle, turn on and inflate, replace cap..''), all of the answerer-evaluator combinations resulted in relatively high scores. 

We also compared the proposed MultiwayPAM with the baseline method of tensor block model (TBM)~\cite{Wang19}. In TBM, we first initialize membership list $\mathcal{M}$ by independently performing $k$-means algorithm on each mode. Then, we repeatedly update $\mathcal{M}$ based on the dissimilarity between the slices of the original and block-wise mean tensors, where the maximum number of iterations was set to $20$. Table \ref{tab:eval} shows the approximation errors for Truthy and Emerton datasets. Here, RMSE-M indicates the root mean squared error (RMSE) between the original score tensor $\mathcal{Y}$ and the medoid tensor $\hat{\mathcal{Y}}$. In regard to TBM method, we defined the medoid of each cluster as the nearest index to the centroid (i.e., mean slice of the cluster). RMSE-C indicates the RMSE between the original score tensor $\mathcal{Y}$ and the centroid tensor $\hat{\mathcal{X}}$, each of whose entry values is a copy of its corresponding centroid entry value. Specifically, it is defined as $\hat{\mathcal{X}} = (\hat{x}_{i_1 \dots i_K})_{i_1 \in [d_1], \dots, i_K \in [d_K]} \in \mathbb{R}^{d_1 \times \dots \times d_K}$, where $\hat{x}_{i_1 \dots i_K} = \frac{1}{|\mathcal{I}_1(i_1)| \dots |\mathcal{I}_K(i_K)|} \sum_{i'_1 \in \mathcal{I}_1(i_1)} \dots \sum_{i'_K \in \mathcal{I}_K(i_K)} y_{i'_1 \dots i'_K}$ for each $(i_1, \dots, i_K)$. Here, $\mathcal{I}_k(i)$ is a set of indices belonging to the cluster of the $i$th index in the $k$th mode. From Table \ref{tab:eval}, we see that the proposed MultiwayPAM achieved a smaller approximation error than TBM in terms of RMSE-M, while its performance was slightly worse than that of TBM in terms of RMSE-C for Truthy dataset.


\begin{table*}[t]
\caption{Members of each cluster for Truthy dataset. Each number indicates the row index in the dataset, and $\bigstar$ indicates the medoid}
\label{tab:cluster_d1}
\centering
\begin{tabular}{|>{\centering}p{8mm}|>{\centering}p{10mm}|p{120mm}|} \hline
Mode & Cluster & Member \\ \hline
\multirow{5}{*}{1} & 1 & Q942, Q887, $\bigstar$Q383, Q738, Q472, Q776, Q32 \\ \cline{2-3}
 & 2 & $\bigstar$Q2, Q46, Q1013, Q409 \\ \cline{2-3}
 & 3 & $\bigstar$Q76, Q659, Q86, Q938, Q104, Q625, Q250, Q78, Q790, Q444, Q636, Q56, Q502, Q268 \\ \cline{2-3}
 & 4 & Q722, Q453, Q840, Q631, Q639, Q562, $\bigstar$Q203, Q353, Q403 \\ \cline{2-3}
 & 5 & Q252, Q241, Q28, Q490, $\bigstar$Q527, Q857, Q123, Q924, Q1004, Q517, Q801, Q15, Q910, Q749, Q142, Q918 \\ \hline
\multirow{5}{*}{2} & 1 & P77935, P134623, $\bigstar$P78249, P110909 \\ \cline{2-3}
 & 2 & P117288, P69220, P40053, P158496, $\bigstar$P176240, P41206, P108401 \\ \cline{2-3}
 & 3 & P12502, P58227, P131931, P173065, P184000, P98003, P118341, P54607, $\bigstar$P73220, P100527, P75760 \\ \cline{2-3}
 & 4 & P51028, P31025, P121024, P94924, P33953, $\bigstar$P101666, P58649, P162401, P110459 \\ \cline{2-3}
 & 5 & P121727, P170620, P57226, P49933, P2386, P143699, $\bigstar$P73570, P80717, P134932, P6172, P143683, P93924, P69074, P113299, P9510, P74236, P82114, P87663, P86932 \\ \hline
\multirow{5}{*}{3} & 1 & P31025, $\bigstar$P49933 \\ \cline{2-3}
 & 2 & P77935, $\bigstar$P69220, P184000, P40053, P78249 \\ \cline{2-3}
 & 3 & $\bigstar$P117288 \\ \cline{2-3}
 & 4 & P58227, $\bigstar$P2386, P9510, P87663, P73220 \\ \cline{2-3}
 & 5 & P51028, P134623, P121727, P121024, P170620, P94924, P57226, P12502, P143699, P131931, $\bigstar$P173065, P73570, P98003, P118341, P80717, P134932, P6172, P33953, P143683, P54607, P93924, P69074, P113299, P101666, P74236, P82114, P158496, P176240, P86932, P58649, P110909, P100527, P162401, P41206, P75760, P108401, P110459 \\ \hline
\end{tabular}
\end{table*}
%
\begin{table*}[t]
\caption{Labels of the medoids for Truthy dataset}
\label{tab:medoid_d1}
\centering
\begin{tabular}{|>{\centering}p{12mm}|p{125mm}|} \hline
Label & Content \\ \hline
Q383 & Can you sense the pressure changes in your environment? \\ \hline
Q2 & Do you possess the ability to navigate or move within a physical environment? \\ \hline
Q76 & What does the sound of a crackling fire feel like to you? \\ \hline
Q203 & Will the world face overpopulation problems in the future due to unchecked human population growth? \\ \hline
Q527 & Do you need to drink exactly eight glasses of water daily to maintain good health? \\ \hline
P78249 & A cynical senior citizen who distrusts corporate practices and lately has been considering options for assisted living facilities for a close friend \\ \hline
P176240 & An author of contemporary American literature who engages in discussions about the evolution of American writing styles \\ \hline
P73220 & a Sri Lankan legal advisor specialized in immigration law \\ \hline
P101666 & An up-and-coming R\&B artist inspired by H.E.R \\ \hline
P73570 & A science fair judge who evaluates projects and provides feedback to students \\ \hline
P49933 & A nurse who supports the cadet's aspiration but worries about the danger involved in a military career \\ \hline
P69220 & The mayor of the community, who recognizes the importance of investing in maternal healthcare \\ \hline
P117288 & A registered dietitian who specializes in sports nutrition and can recommend the most suitable snacks for athletes \\ \hline
P2386 & A young mother living in Colorado with a four-year-old child \\ \hline
P173065 & A software engineer who is familiar with React and web components, but does not know much about SkateJS \\ \hline
\end{tabular}
\end{table*}
\begin{table*}[t]
\caption{Members of each cluster for Emerton dataset}
\label{tab:cluster_d2}
\centering
\begin{tabular}{|>{\centering}p{8mm}|>{\centering}p{10mm}|p{120mm}|} \hline
Mode & Cluster & Member \\ \hline
\multirow{5}{*}{1} & 1 & $\bigstar$Q16 \\ \cline{2-3}
 & 2 & Q4011, $\bigstar$Q4981, Q883, Q5402 \\ \cline{2-3}
 & 3 & $\bigstar$Q2000 \\ \cline{2-3}
 & 4 & Q43, Q1920, Q473, $\bigstar$Q2183, Q5169, Q1419, Q4335, Q1737, Q3866, Q40, Q2440, Q5155, Q5123, Q3218, Q3885, Q2256, Q3765 \\ \cline{2-3}
 & 5 & Q5304, Q4047, Q1594, Q4858, Q2785, $\bigstar$Q4376, Q135, Q4182, Q1458, Q3109, Q4319, Q777, Q3833, Q5064, Q3014, Q1454, Q2326, Q3660, Q609, Q2266, Q2967, Q4112, Q4423, Q3028, Q4521, Q3425, Q2286 \\ \hline
\multirow{5}{*}{2} & 1 & P87836, P173373, P182397, $\bigstar$P52793, P41948, P87539, P78208, P107465, P170459, P140471, P66072, P6192, P1654, P81177, P16714, P38363, P172511, P188434, P49988, P182013, P85812 \\ \cline{2-3}
 & 2 & P156543, P86290, P189700, P106245, P189518, $\bigstar$P189606, P139819, P64043 \\ \cline{2-3}
 & 3 & P31950, P147135, P143044, P136894, P88376, $\bigstar$P69025, P189925, P74639 \\ \cline{2-3}
 & 4 & P133846, P188759, P100252, P112451, $\bigstar$P29365, P187221 \\ \cline{2-3}
 & 5 & P67461, $\bigstar$P179289, P185670, P60199, P15731, P59988, P127673 \\ \hline
\multirow{5}{*}{3} & 1 & $\bigstar$P38363, P49988 \\ \cline{2-3}
 & 2 & P52793, P41948, P133846, $\bigstar$P31950, P16714, P172511, P74639, P85812 \\ \cline{2-3}
 & 3 & P173373, P182397, P87539, P60199, P86290, P170459, P100252, $\bigstar$P15731, P106245, P1654, P81177, P136894, P127673, P64043, P189925 \\ \cline{2-3}
 & 4 & P87836, P67461, P179289, P185670, $\bigstar$P6192, P143044, P88376, P69025 \\ \cline{2-3}
 & 5 & P156543, P78208, P107465, P188759, P140471, P147135, $\bigstar$P189700, P66072, P189518, P59988, P112451, P189606, P139819, P188434, P29365, P187221, P182013 \\ \hline
\end{tabular}
\end{table*}
%
\begin{table*}[t]
\caption{Labels of the medoids for Emerton dataset}
\label{tab:medoid_d2}
\centering
\begin{tabular}{|>{\centering}p{12mm}|p{125mm}|} \hline
Label & Content \\ \hline
Q16 & Imagine a question and stream-of-consciousness explanation for which this is the answer: Sentence B \\ \hline
Q4981 & Given the task definition and input, reply with output. A text is given in Tamil. Translate it from the Tamil language to the Hindi language. The translation must not omit or add information to the original sentence. $\dots$ \\ \hline
Q2000 & After the death of Spitamenes and his marriage to Roxana (Roshanak in Bactrian) to cement relations with his new satrapies, Alexander turned to the Indian subcontinent. He invited the chieftains of the former satrapy of Gandhara, in the north of what is now Pakistan, to come to him and submit to his authority. $\dots$ Choose your answer: According to the above paragraph, the correct answer to the question "What did Alexander do following the death of his wife Roxana?" is "He invited the chieftains of the former satrapy of Gandhara to come to him and submit to his authority"?  OPTIONS: (a). no (b). yes \\ \hline
Q2183 & Continue writing the next sentence in this paragraph:  How to calculate amount of unemployment in ohio  Meet the criteria for unemployment insurance.  There are a certain number of factors that determine whether or not you are eligible for unemployment compensation. If the following applies to you, then you are considered unemployed and are eligible for unemployment insurance and benefits. \\ \hline
Q4376 & Objective: How to inflate a car tire.  Which of the following solutions is more sound in terms of naive physics reasoning?  Options are: A). Remove valve stem cap, attach air nozzle, turn on and inflate, don't replace cap.. B). Remove valve stem cap, attach air nozzle, turn on and inflate, replace cap.. \\ \hline
P52793 & A descendant of a prominent Native American tribe leader seeking to trace their family lineage \\ \hline
P189606 & A long-time fan of Joss Stone whose favorite holiday is Christmas \\ \hline
P69025 & A successful mathematician who attributes their career to the solid foundation in math provided by this physics teacher \\ \hline
P29365 & A patient psychology student who often models for their friend's paintings \\ \hline
P179289 & A perfectionist art student, struggling to embrace a more fluid and open-ended approach in their creation \\ \hline
P38363 & A securities lawyer who believes that corporate law favors large corporations and perpetuates income inequality \\ \hline
P31950 & An e-commerce entrepreneur who believes in bridging the gap between online and offline shopping by offering click-and-collect services \\ \hline
P15731 & A blogger and book reviewer who recommends uplifting and thought-provoking novels to the patient \\ \hline
P6192 & A casual listener of classical music who avows an affinity for Beethoven and Mozart \\ \hline
P189700 & A chauffeur responsible for transporting the philanthropist to charity events and meetings with beneficiaries \\ \hline
\end{tabular}
\end{table*}

\begin{table}[t]
\caption{Approximation errors for Truthy and Emerton datasets}
\label{tab:eval}
\centering
\begin{tabular}{ccccc}
\toprule
& \multicolumn{2}{c}{Truthy dataset} & \multicolumn{2}{c}{Emerton dataset} \\ \cmidrule{2-5}
 & MPAM & TBM & MPAM & TBM \\ \midrule
RMSE-M & 0.704 & 0.824 & 0.530 & 0.574 \\
RMSE-C & 0.666 & 0.651 & 0.509 & 0.529 \\ \bottomrule
\end{tabular}
\end{table}


\subsection{Missing Value Prediction based on Estimated Block Structure}

We also compared missing value prediction performance of the proposed MultiwayPAM and TBM. Let $p \in [0, 1]$ be a ratio of missing values in a given score tensor. In the experiment, we tried $p \in \{0.1, 0.3, 0.5, 0.7, 0.9\}$. We randomly selected entries of the score tensor at a ratio $p$ and treated them as missing values. Specifically, the values of selected entries were replaced with (1) the constant $2.5$ (i.e., the average of the possible scores) or (2) scores randomly selected from $\{1, 2, 3, 4\}$. We applied the proposed MultiwayPAM and TBM to the score tensor after such replacement operation. As for the number of clusters, we tried $\bm{c} = [k, k, k]^{\mathsf{T}}$ for $k \in \{3, 5, 7\}$. Based on the estimated block structure, we predicted each missing value by the medoid or centroid score. Finally, we calculated the RMSE between the predicted values and the original values before replacement. For each setting, we tried such replacement and prediction five times and evaluated the mean and standard deviation of the RMSE scores.

\begin{figure*}[t]
\centering
\includegraphics[width=\textwidth]{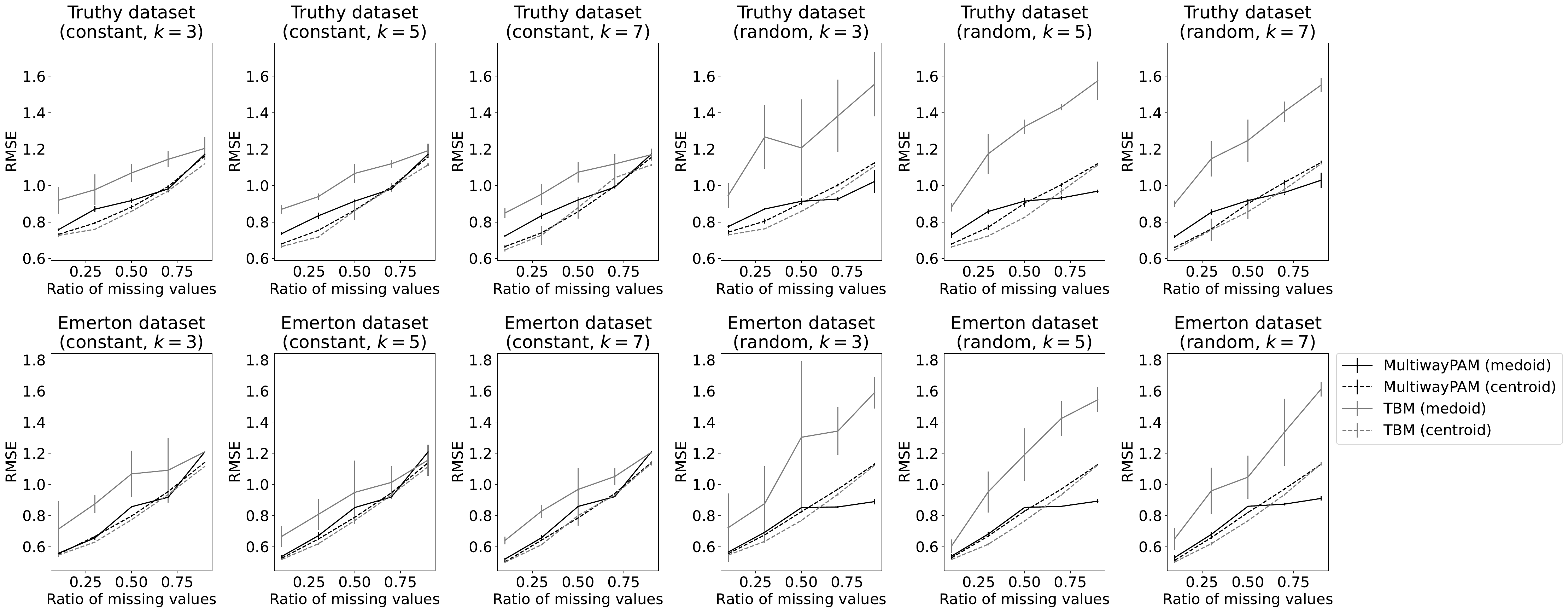}
\caption{Missing value prediction performance of the proposed MultiwayPAM and TBM for Truthy and Emerton datasets. ``Constant'' and ``random'' indicate the types of values to replace missing values, and $k$ is the number of clusters in each mode. In the legend, (medoid) and (centroid) indicate the methods for predicting missing values based on the estimated block structure}\vspace{3mm}
\label{fig:predict}
\end{figure*}

Figure \ref{fig:predict} shows the RMSE scores of missing value prediction performance for each setting. While the proposed method improves the interpretability of the estimated block structure by providing medoids, in most settings where $p \leq 0.5$, the centroid-based prediction using TBM was the most accurate. On the other hand, the accuracy gap between the centroid-based and medoid-based predictions was smaller when using the proposed MultiwayPAM than when using TBM.

\section{Conclusion}

In this paper, we proposed a new tensor clustering method MultiwayPAM for simultaneously estimating the block structure and the medoids for each mode of a given LLM-as-a-Judge score tensor. By extending PAM algorithm for vector data clustering, where we repeatedly update the cluster memberships and the medoids, we can obtain the local optimal solution. We applied the proposed MultiwayPAM to the score tensors for two practical datasets and showed its effectiveness. In MultiwayPAM, we assume that the cluster size vector $\bm{c}$ is given. However, in some cases, the appropriate cluster sizes would be unknown in advance. Therefore, it is an important topic in future study to determine the appropriate number of blocks for a given data tensor, along with the cluster memberships and the medoids. Furthermore, although the medoids obtained by MultiwayPAM locally minimize the dissimilarity between the original and medoid tensors, they are not necessarily semantically close to the other indices in the same clusters. Practically, it would be an important future work to develop a tensor clustering method for selecting medoids by considering both approximation of the original data tensor and intra-cluster semantic similarity.


\bibliographystyle{unsrt}
\bibliography{cas-refs}

\end{document}